\documentclass[sigconf]{acmart}
\usepackage{multirow}
\usepackage{bm}
\usepackage{graphicx}
\usepackage{amsmath}
\usepackage{soul}
\usepackage{color}
\usepackage{amsfonts}
\usepackage[ruled,linesnumbered]{algorithm2e}
\definecolor{beaublue}{rgb}{0.74, 0.83, 0.9}

\AtBeginDocument{%
  \providecommand\BibTeX{{%
    \normalfont B\kern-0.5em{\scshape i\kern-0.25em b}\kern-0.8em\TeX}}}

\copyrightyear{2021} 
\acmYear{2021} 
\setcopyright{acmlicensed}\acmConference[CIKM '21]{Proceedings of the 30th ACM International Conference on Information and Knowledge Management}{November 1--5, 2021}{Virtual Event, QLD, Australia}
\acmBooktitle{Proceedings of the 30th ACM International Conference on Information and Knowledge Management (CIKM '21), November 1--5, 2021, Virtual Event, QLD, Australia}
\acmPrice{15.00}
\acmDOI{10.1145/3459637.3481946}
\acmISBN{978-1-4503-8446-9/21/11}

\begin{document}

\title{QUEACO: Borrowing Treasures from Weakly-labeled Behavior Data for Query Attribute Value Extraction}
\author{Danqing Zhang*$^1$, Zheng Li*$^1$, Tianyu Cao$^1$, Chen Luo$^1$, Tony Wu$^1$, Hanqing Lu$^1$, \\Yiwei Song$^1$, Bing Yin$^1$, Tuo Zhao$^2$, Qiang Yang$^3$}
\affiliation{
	\institution{$^1$Amazon.com Inc, $^2$Georgia Institute of Technology, $^3$Hong Kong University of Science and Technology}
	\institution{$^1$\{danqinz,amzzhe,caoty,cheluo,tonywu,luhanqin,ywsong,alexbyin\}@amazon.com,\\ $^2$tourzhao@gatech.edu, $^3$qyang@cse.ust.hk}
	\country{}
}

\thanks{*Authors contributed equally to this paper.}

\begin{abstract}
We study the problem of query attribute value extraction, which aims to identify named entities from user queries as diverse surface form attribute values and afterward transform them into formally canonical forms. Such a problem consists of two phases: {named entity recognition (NER)} and {attribute value normalization (AVN)}. However, existing works only focus on the NER phase but neglect equally important AVN. To bridge this gap, this paper proposes a unified query attribute value extraction system in e-commerce search named QUEACO, which involves both two phases. Moreover, by leveraging large-scale weakly-labeled behavior data, we further improve the extraction performance with less supervision cost. Specifically, for the NER phase, QUEACO adopts a novel teacher-student network, where a teacher network that is trained on the strongly-labeled data generates pseudo-labels to refine the weakly-labeled data for training a student network. Meanwhile, the teacher network can be dynamically adapted by the feedback of the student's performance on strongly-labeled data to maximally denoise the noisy supervisions from the weak labels. For the AVN phase, we also leverage the weakly-labeled query-to-attribute behavior data to normalize surface form attribute values from queries into canonical forms from products. Extensive experiments on a real-world large-scale E-commerce dataset demonstrate the effectiveness of QUEACO.
\end{abstract}



\maketitle
\section{Introduction}

Query attribute value extraction is the joint task of detecting named entities in the search queries as the diverse surface form attribute values and normalizing them into a canonical form to avoid misspelling and abbreviation problems. These two sub-tasks are typically called named entity recognition (NER)~\cite{chiu2016named} and attribute value normalization (AVN) ~\cite{putthividhya2011bootstrapped}. 

\begin{figure}[htb!]
\centering
\vspace{-3mm}
\includegraphics[width=0.9\linewidth]{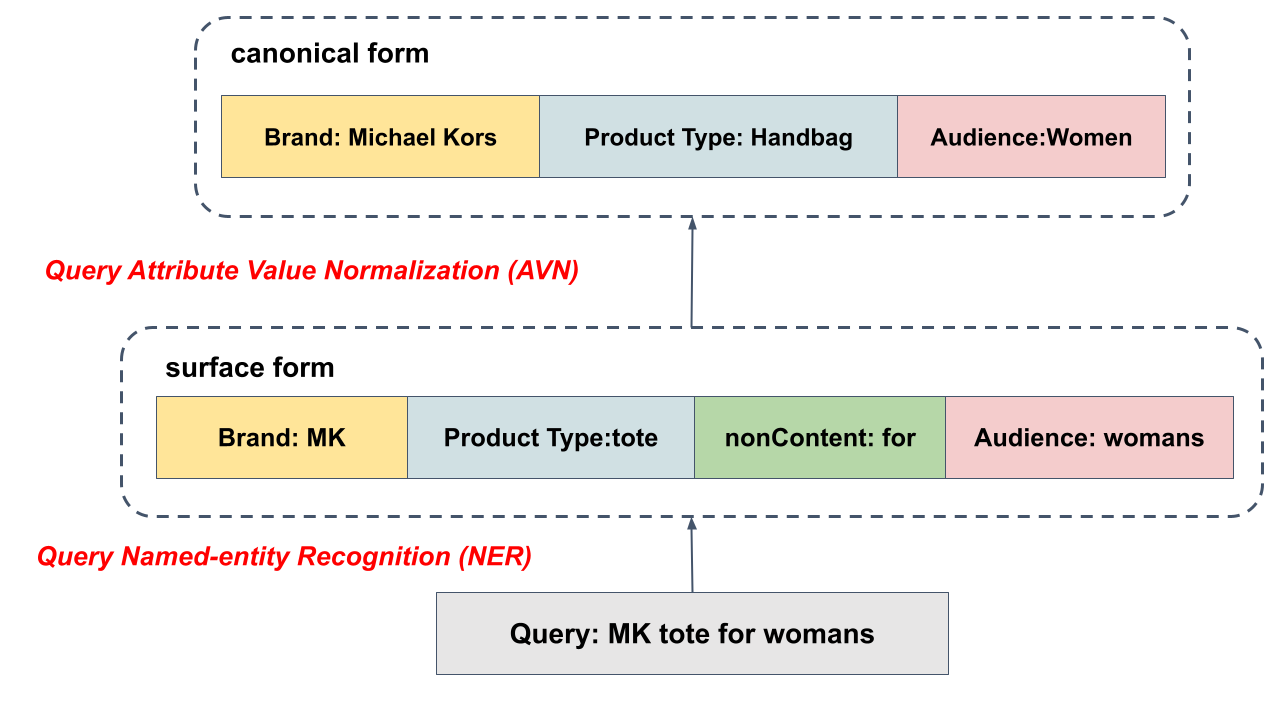}
\vspace{-5mm}
\caption{The ideal product attribute extraction pipeline. }
\label{fig:framework}
\vspace{-3mm}
\end{figure}

\begin{table*}[h]
    \centering
    \fontsize{5.0}{6}\selectfont
    \resizebox{2\columnwidth}{!}
    {%
    \begin{tabular}{c | l | l| l  } 
        \hline
        Case\# & Query \& Ground-truth Labels & Clicked Product Attribute Values & Weak Labels  \\ \hline
        1 &  \textbf{[}{\color{orange}\textit{lg}}\textbf{]}\textbf{[}{\color{blue}\textit{smart tv}}\textbf{]}\textbf{[}{\color{green}\textit{32}}\textbf{]} & {\color{orange}\textit{lg}}, {\color{green}\textit{32-inch}}, {\color{blue}\textit{television}}  & \textbf{[}{\color{orange}\textit{lg}}\textbf{]} \textit{smart tv 32} 
        \\\hline
        2 & \textbf{[}{\color{magenta}\textit{womans}}\textbf{]}\textbf{[}{\color{blue}\textit{socks}}\textbf{]} & {\color{magenta}\textit{women}}, {\color{blue}\textit{socks}} &  \textit{womans} \textbf{[}{\color{blue}\textit{socks}}\textbf{]}
        \\ \hline
        3 & \textbf{[}{\color{orange}\textit{braun}}\textbf{]}\textbf{[}{\color{cyan}\textit{7 series}}\textbf{]}\textbf{[}{\color{blue}\textit{shaver}}\textbf{]} & {\color{orange}\textit{braun}}, {\color{cyan}\textit{series 7}}, {\color{blue}\textit{electric shaver}}  & \textbf{[}{\color{orange}\textit{braun}}\textbf{]} \textit{7 series shaver}
        \\ \hline
        4 & \textbf{[}{\color{orange}\textit{trixie}}\textbf{]} \textbf{[}{\color{blue}\textit{cat litter tray bags}}\textbf{]}\textbf{[}{\color{green}\textit{46 x 59}}\textbf{]}\textbf{[}{\color{green}\textit{10 pack}}\textbf{]} & {\color{orange}\textit{Trixie}}, {\color{blue}\textit{waste bag}}, {\color{green}\textit{46 × 59 cm}} & \textbf{[}{\color{orange}\textit{trixie}}\textbf{]} \textit{cat litter tray bags 46 x 59 10 pack}\\
        \hline
    \end{tabular}
    }
    \caption{Ground-truth labels and noisy weakly-labels for query NER examples based on the behavior data from the product side. We use colors to denote the entity type and use brackets to indicate the entity boundary.
    Entity labels: {{\color{orange}\textrm{Brand}}}, 
    {{\color{cyan}\textrm{ProductLine}}}, 
    {{\color{green}\textrm{Size}}}, 
    {{\color{blue}\textrm{ProductType}}},
    {{\color{magenta}\textrm{Audience}}}.
    }
    \label{tab:ws_example}
    \vspace{-2mm}
\end{table*}

As shown in Figure~\ref{fig:framework}, we illustrate the process of the ideal query attribute value extraction. When a user enters the query ``{\it MK tote for womans}'', we firstly use a NER model to identify the entity type ``brand'' for "{\it MK}", ``product type'' for "{\it tote}", and ``audience'' for ``{\it womans}''. These extracted named entities are in the informal surface form of attribute values. However, such an informal surface is not accordant with the products indexed with canonical form attribute values in the formal written style. Specifically, ``{\it MK}'' is an abbreviation of brand ``{\it Michael Kors}'', ``{\it tote}'' is a hyponym of the product type "{\it handbag}", and ``{\it womans}'' contains a spelling error. This misalignment poses tremendous challenges to the product search engine to retrieve relevant product items that users really prefer. Therefore, the AVN module is equally important to transform the surface form for each attribute value into the canonical form, i.e., ``{\it MK}'' to ``{\it Michael Kors}'', ``{\it tote}'' to ``{\it handbag}'' and ``{\it womans}'' to ``{\it women}''. In the E-commerce domain, extracting these attributes values from queries is critical to a wide variety of product search applications, such as product retrieval~\cite{cheng2020end} and ranking~\cite{wen2019building}, and query rewriting~\cite{guisado2016enrich}.

Unfortunately, existing works only focus on the surface form attribute value extraction based on NER while ignoring the canonical form transformation, which is impractical in the realistic scenarios~\cite{kozareva2016recognizing, cheng2020end,wen2019building,cowan2015named}. To bridge this gap, this paper proposes a unified query attribute value extraction system that involves both phases. By borrowing treasures from large-scale weakly-labeled behavior data to mitigate the supervision cost, we further improve the extraction performance. Considering the first NER stage, recent advances in deep learning models (e.g., Bi-LSTM+CRF) have achieved promising results~\cite{huang2015bidirectional,raganato2017neural}. However, they highly rely on massive labeled data, where manual labeling for token-level labels is particularly costly and labor-intensive. To alleviate the issue in E-commerce, prior studies \cite{kozareva2016recognizing,cheng2020end,wen2019building} resort to leveraging large-scale behavior data from the product side as the weak supervision for queries based on some simple string match strategies. 
Nonetheless, these weakly-supervised labels contain enormous noises due to the partial or incomplete token labels based on the exact string matching. For example, as shown in Table~\ref{tab:ws_example} case\#1, when we use the attribute values of top-clicked product ``{LG 32-inch television}'', i.e., ``brand'' for ``{\it LG}'', ``size'' for ``{\it 32-inch}'', ``product type'' for ``{\it television}''as the weak supervision to match the query ``{\it lg smart tv 32}'', it can only generate the label ``brand''  for ``{\it lg}'', concealing useful knowledge for the unannotated tokens.
For this reason, weak supervision-based methods~\cite{shang2018learning,cheng2020end} usually perform very poorly, even worse after powerful pre-trained language models (PLMs) (e.g., BERT~\cite{devlin2019bert} ) are introduced since PLMs are much easier to fit noises. To address the issue, we consider a more reliable regime, which further includes some strongly-labeled human annotated data to denoise the weak labels from the distant supervision. As such, the NER model can be improved by making more effective use of both the large-scale weakly-labeled behavior data and the strongly-labeled human-annotated data.

As for the second AVN phase, customers tend to use diverse surface forms to mention each attribute value in search queries due to the misspellings, spelling variants, or abbreviations. This circumstance occurs frequently in user queries and product titles of e-commerce. For example, eBay has noted that 20\% product titles in the clothing and shoes category involve such surface form brand~\cite{putthividhya2011bootstrapped}. Thus, normalizing these surface form attribute values derived from the NER signals to a single normalized attribute value is critical. It is usually ignored by existing works~\cite{kozareva2016recognizing, cheng2020end,wen2019building,cowan2015named}. To mitigate human annotating efforts, weakly-labeled behavior data can also contribute to the AVN. For example, ``{\it MK tote for womans}'' mentioning the brand ``{\it MK}'' leads to the click of product items associated with the brand ``{\it Michael Kors}''. We can reasonably infer a strong connection between the surface form value ``{\it MK}'' and the canonical form value ``{\it Michael Kors}'' if this association occurs in many queries.

Motivated by those, we propose a unified \textbf{QUE}ry \textbf{A}ttribute Value Extraction in E\textbf{CO}mmerce (\textbf{QUEACO}) framework that efficiently utilizes the large-scale weakly-labeled behavior data for both the query NER and AVN. For query NER, QUEACO leverages the strongly-labeled data to denoise the weakly-labeled data based on a novel teacher-student network, where a teacher network trained on the strongly-labeled data generates pseudo-labels to refine the weakly-labeled data for teaching a student network. Unlike the classic teacher-student networks that can only produce pseudo-labels from a fixed teacher, our pseudo-labeling process from the teacher is continuously and dynamically adapted by the feedback of the student's performance on the strongly-labeled data. This encourages the teacher network to generate better pseudo-labels to teach the student, maximally mitigating the error propagation from the noisy weak labels. For query AVN, we utilize the weakly-labeled query-to-attribute behavior data and QUEACO NER predictions to model the associations between the surface form and canonical form attribute values. As such, the surface form attribute values from queries can be normalized to the most relevant canonical form attribute values from the products. Empirically, extensive experiments on a real-world large-scale E-commerce dataset demonstrate that QUEACO NER can significantly outperform the state-of-the-art semi-supervised and weakly-supervised methods. Moreover, we qualitatively show the effectiveness and the necessity of QUEACO AVN. 

Our contributions can be summarized as follows: (1) To the best of our knowledge, our work is the first attempt to propose a unified query attribute value extraction system in E-commerce, involving both the query NER and AVN. QUEACO can automatically identify product-related attributes from user queries and transform them into canonical forms, by leveraging weak supervisions from large-scale behavior data; (2) Our QUEACO NER is also the first work that efficiently utilizes both human-annotated strongly-labeled data and large-scale weakly-labeled data from the query-product click graph. Moreover, the proposed QUEACO NER model can significantly outperform the existing state-of-the-art baselines; (3) We propose the QUEACO AVN module that uses aggregated query to attribute behavioral data to build the connections among queries, surface form attribute value, and canonical form value. The proposed QUEACO AVN module can effectively normalize the surface form values with spelling errors, spelling variants, and abbreviations problems.

\section{Preliminaries}

In this section, we introduce some preliminaries before detailing the proposed QUEACO framework, including the problem formulation and the query NER base model.

\subsection{Problem Formulation}
\subsubsection{QUEACO Named Entity Recognition}
We firstly introduce the task definition for the QUEACO NER.

\noindent \textbf{NER} \indent Given a user input query ${\mathbf{X_{i}}\!=\![x_{1}, x_{2},...,x_{M}]}$ with $M$ tokens, the goal of NER is to predict a tag sequence $\small{{\mathbf{Y_{i}}\!=\![y_{1}, y_{2},...,y_{M}]}}$. We use the BIO ~\cite{li2012joint} tagging strategy. Specifically, the first token of an entity mention with each entity type $C_o\in C$ ($C$ is the entity type set) is labeled as $\mathrm{B\!-\!C_o}$; the remaining tokens inside that entity mention are labeled as $\mathrm{I\!-\!C_o}$; and the non-entity tokens are labeled as $\mathrm{O}$.

\noindent \textbf{Strongly-Labeled and Large Weakly-Labeled Setting} \indent 
For our query NER, we have two types of data: 1) strongly-labeled  data $D_{l}\!=\!\{(\mathbf{X}_{i}^{l}, \mathbf{Y}_{i}^{l})\}_{i=1}^{N_{l}}$, which is manually annotated by human annotators; 2) large-scale weakly-labeled data $D_{w}\!=\!\{(\mathbf{X}_{i}^{w}, \mathbf{Y}_{i}^{w})\}_{i=1}^{N_{w}}$, where $N_{l}\!\ll\!N_{w}$. The goal is to borrow treasures from large-scale noisy weakly-labeled data to further enhance a supervised NER model trained on the strongly-labeled data.

\subsubsection{QUEACO Attribute Value Normalization}\label{avn_notations}
For each query \\${\mathbf{X_{i}}\!=\![x_1, x_2,...,x_M]}$ with $M$ tokens, QUEACO NER predicts a tag sequence ${\mathbf{\widetilde{Y}_{i}}\!=\![\Tilde{y_1}, \Tilde{y_2},...,\Tilde{y_M}]}$. Given a entity type $C_o \in C$ (e.g., brand) and the NER prediction $\mathbf{\widetilde{Y}_{i}}$, we can extract the query term $\mathbf{X}^{C_o}_{i}$ as the surface form attribute value for the entity type $C_o$. Assume that we have a diverse set of canonical form product attribute values $\mathbf{V}$ for the entity type $C_o$. For each canonical form attribute value $v\in \mathbf{V}$, we can define the relevance given the query $\mathbf{X}_{i}$ as

\begin{equation*}
P(C_o=v|\mathbf{X}_{i}) =\frac{\sum_{d \in D} n(d,\mathbf{X}_{i}) \vmathbb{1}(d_{C_{o}} = v) }{\sum_{d \in D} n(d,\mathbf{X}_{i})}
\end{equation*}
where $n(d,\mathbf{X}_{i})$ is the number of total clicks on the product $d$ of the the searches using query $\mathbf{X}_{i}$ in a period of time, such as one month. And $D$ is the set of all products. $\vmathbb{1}(d_{C_{o}} = c)$ indicates whether the product $d$ is indexed with the value $c$ for the entity type $C_o$. In a nutshell, we quantify the query-attribute relevance using the query-product relevance and the product-attribute membership. The query-product relevance is measured by number of clicks in the query logs, which can be viewed as the implicit feedback from customers. Finally, we can get the most relevant attribute value of the entity type $C_o$ by $\arg\max {P(C_o=\mathbf{V}|\mathbf{X}_{i})}$ as the normalized canonical form for the surface form attribute value $\mathbf{X}^{C_o}_{i}$.

\subsection{Query NER Base Model}\label{model_architecture_benchmark}
The recent emergence of the pre-trained language models (PLMs) such BERT~\cite{devlin2019bert} has achieved superior performance on a variety of public NER datasets. However, existing query NER works~\cite{cheng2020end,wen2019building,cowan2015named,kozareva2016recognizing} still rely on the shallow deep learning models (e.g., BiLSTM-CRF) while not equipping with the powerful PLMs.

\textbf{Why PLMs are not deployed for existing query NER works?} Due to labeled data scarcity in user queries, previous query NER works can only rely on the noisy distant supervision data for model training. In such a condition, using the powerful mPLMs as the encoder has even worse performance than a shallow Bi-LSTM for the query NER~\cite{cheng2020end}. ~\citet{liang2020bond} have found that the PLM-based NER models are easier to overfit the noises from the distant labels and forget the general knowledge from the pre-training stage. On the other hand, distant supervision based methods for NER~\cite{shang2018learning,cheng2020end} usually underperform, which cannot meet the high performance requirement for query NER used by various downstream applications in product search like retrieval and ranking. To tackle the issue, we target a different query NER setting, which leverages some strongly-labeled human-annotated data to train a more reliable PLM-based NER model and uses the weakly-labeled data from the distant supervision to further improve the model performance. To meet the strict latency constraint, we choose DistilmBERT~\cite{Sanh2019DistilBERTAD} as the base NER model and we do not add the CRF layer. 

\begin{figure*}[t]
\centering
\includegraphics[width=1\linewidth]{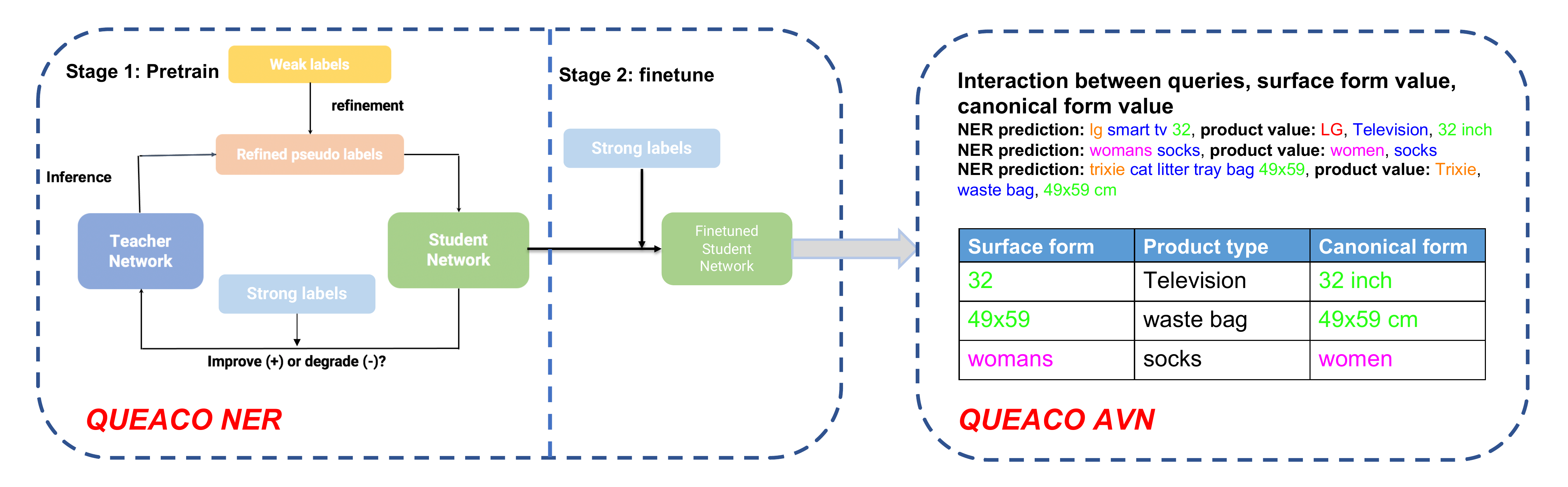}
\caption{An overview of the proposed framework QUEACO, showing how weakly behavioral data contributes to the two inter-dependent stages of QUACO. Entity labels: {{\color{orange}\textrm{Brand}}}, 
    {{\color{cyan}\textrm{ProductLine}}}, 
    {{\color{green}\textrm{Size}}}, 
    {{\color{blue}\textrm{ProductType}}},
    {{\color{purple}\textrm{namedPersonGroup}}},
    {{\color{red}\textrm{Color}}},
    {{\color{magenta}\textrm{Audience}}}.}
\label{fig:QUEACO_framework}
\vspace{-3mm}
\end{figure*}

\section{QUEACO}

In this section, we firstly give an overview of how weakly-labeled behavior data contributes to both the query NER and AVN and then detail the two components for QUEACO, respectively.

\subsection{Overview}
Figure~\ref{fig:QUEACO_framework} shows an overview of QUEACO. At a high level, QUEACO leverages weakly-labeled behavior data for both the query NER and AVN. For QUEACO NER, we have the strongly-labeled data and the large-scale weakly-labeled data for training. Specifically, the QUEACO NER has two stages: the weak supervision pretraining stage and the finetuning stage. 1) In the pretraining stage, we adopt a novel teacher-student network where the teacher network is dynamically adapted based on the feedback from the student network. The goal is to encourage the teacher network to generate better pseudo labels to refine the weakly-labeled data for improving the student network's performance. 2) After the pretraining stage, we continue to finetune the student network on the strongly-labeled data as the final model. For QUEACO AVN, we extract the surface form attribute values based on the NER predictions and leverage the weakly-labeled query-to-attribute behavior data to transform them into the canonical forms.

\subsection{QUEACO Named Entity Recognition}\label{query_ner_model}

\subsubsection{Model architecture}
\noindent \textbf{Teacher-Student Network} 
Before introducing the QUEACO NER model, we give some preliminary of the teacher-student network of self-training~\cite{lee2013pseudo,yarowsky1995unsupervised}. Self-training stands out among semi-supervised learning approaches, in which a teacher model produces pseudo-labels for unlabeled samples, and a student model learns from these samples with generated pseudo-labels. We give the mathematical formulation of self-training in the context of NER. Let $T$ and $S$ respectively be the teacher and student network, parameterized by $\bm{\theta}_{T}$ and $\bm{\theta}_{S}$. We use $f(\mathbf{X};\bm{\theta}_{T})$ and $f(\mathbf{X};\bm{\theta}_{S})$ denote the NER predictions of the query $\mathbf{X}$ for the teacher and student, respectively. $f(\mathbf{X};\bm{\theta}_{T})$ can be either soft or converted to hard pseudo labels. Then the knowledge transfer is usually achieved by minimizing the consistency loss between the two predicted distributions from the teacher and the student: $\mathcal{L}(f(\mathbf{X};\bm{\theta}_{T}), f(\mathbf{X};\bm{\theta}_{S}))$. \\

\noindent \textbf{Pseudo \& Weak Label Refinement} 
Weakly-labeled data suffers from severe incompleteness that the overall span recall is usually very low. Therefore, it is natural to use self-training to annotate the missing labels of the weakly-labeled data. The pseudo labels make up the missing tags for the weak labels, and meanwhile weak labels can provide high precision tags to restrict pseudo labels.

For each weakly-labeled sample $\mathbf{X}_{i}^{w} \!=\![ {x_{1}^{w}, x_{2}^{w}},...,x_{M}^{w}]$, we convert the soft predictions of the teacher network into the hard pseudo labels, i.e.,  $\mathbf{Y}_{i}^{p} = \mathop{\arg\max} f(\mathbf{{X}}_{i}^w; \bm{\theta}_{T}) \!=\![y_{1}^{p},y_{2}^{p},...,y_{M}^{p}]$. Additionally, we have weak labels $\mathbf{Y}_{i}^{w} =  \!=\![y_{1}^{w},y_{2}^{w},...,y_{M}^{w}]$ that partially annotate the samples, which can be used to further refine the pseudo labels.  We maintain the weak labels of the entity tokens and replace the weak labels of the no entity tokens with the pseudo labels. Then the refined pseudo labels $\mathbf{Y}_{i}^{r} =  \!=\![y_{1}^{r},y_{2}^{r},...,y_{M}^{r}]$ are generated by:
 \[
    y_{j}^{r}= 
\begin{cases}
    y_{j}^{p},& \text{if } y_{j}^{w} = \mathrm{O}\\
    y_{j}^{w},              & \text{otherwise}
\end{cases}
\]

\noindent \textbf{QUEACO Teacher-Student Network}
Prior teacher-student frameworks of self-training rely on rigid teaching strategies, which may hardly produce high-quality pseudo-labels for consecutive and interdependent tokens. This results in progressive drifts on the noisy pseudo-labeled data provided by the teacher (a.k.a the confirmation bias~\cite{arazo2020pseudo}). In QUEACO NER, we propose a novel teacher-student network, where the teacher can be dynamically adapted from the student's feedback to adjust its pseudo-labeling strategies, inspired by \citet{pham2021meta}. Student's feedback is defined as the student's performance on the strongly-labeled data. Formally, we can formulate our teacher-student network as a bi-level optimization problem,

\begin{equation*}
\begin{aligned}
\min_{\bm{\theta}_{T}} \quad & \mathcal{L}_{S,l} (\bm{\theta}_{S}^{t+1}(\bm{\theta}_{T}))\\
\textrm{s.t.} \quad & 
\bm{\theta}_{S}^{t+1}(\bm{\theta}_{T}) =\mathop{\arg\min}_{\bm{\theta_{S}}} \frac{1}{N_{w}}\sum_{i=1}^{N_{w}} \ell(\mathbf{Y}_i^{r}, f(\mathbf{X}_i^{w}; \bm{\theta}_{S}^{t})). \\
\end{aligned}
\end{equation*}

where $\ell$ is the cross-entropy loss. The ultimate goal is to minimize the loss of the student $\bm{\theta}_{S}^{t+1}$ on the strongly-labeled data after learning from the refined pseudo labels $\mathbf{Y}_{i}^{r}$, i.e., $\mathcal{L}_{S,l} (\bm{\theta}_{S}^{t+1}(\bm{\theta}_{T}))$, which is a function of
the teacher's parameters $\bm \theta_{T}$. $f(\mathbf{X}_i^{w}; \bm{\theta}_{S}^{t})$ is the prediction logits of the student network on the weakly-labeled sample $\mathbf{X}_{i}^{w}$. By optimizing the teacher's parameter in light of the student's performance on the strongly-labeled data, the teacher can be adapted to generate better pseudo labels to further improve student's performance. This bi-level optimization problem is extremely complicated, but we can approximate the multi-step $\mathop{\arg\min}_{\bm{\theta}_{S}}$ with one step gradient update of $\bm{\theta}_{S}$. Plugging this into the constrained optimization problem leads to an unconstrained optimization for the teacher network learning. This gives rise to the alternating optimization procedure between the student and the teacher updates.\\

\subsubsection{Model Training}
\noindent \textbf{Student Network} 
The student network is trained with refined pseudo-labeled data $\mathbf{Y}_{i}^{r}$ in order to move closer to the teacher, 
\begin{equation*}
\mathcal{L}_{\mathrm{S}}=\frac{1}{N_{w}}\sum_{i=1}^{N_{w}} \ell(\mathbf{Y}_i^{r}, f(\mathbf{X}_i^{w}; \bm{\theta}_{S}))    
\end{equation*}

We update student network parameter $\bm{\theta}_{S}$ with one step of gradient descent. In our proposed framework, the feedback signal from the student network to the teacher network is the student's performance on the strongly-labeled data. We use the student loss on the strongly-labeled data to measure the performance before the update ($\bm{\theta}_{S}^{t}$) and after the update ($\bm{\theta}_{S}^{t+1}$, learning on the refined pseudo-labeled data),
\begin{equation*}
\small{
\begin{aligned}
\mathcal{L}_{S,\mathrm{l}}^{(t)}=\frac{1}{N_{l}}\sum_{i=1}^{N_{l}} \ell(\mathbf{Y}_i^{l}, f(\mathbf{X}_i^{l}; \bm{\theta}_{S}^{t})),\\
\mathcal{L}_{S,\mathrm{l}}^{(t+1)}=\frac{1}{N_{l}}\sum_{i=1}^{N_{l}} \ell(\mathbf{Y}_i^{l}, f(\mathbf{X}_i^{l}; \bm{\theta}_{S}^{t+1})).
\end{aligned}}
\end{equation*}
The difference between $\mathcal{L}_{S,\mathrm{l}}^{(t)}$ and  $\mathcal{L}_{S,\mathrm{l}}^{(t+1)}$, i.e., $\small{\lambda_{\mathrm{meta}}\!=\!\mathcal{L}_{S,\mathrm{l}}^{(t+1)}\!-\!\mathcal{L}_{S,\mathrm{l}}^{(t)}}$, can be used as the feedback to meta-optimize the teacher network towards the direction that generates better pseudo labels. If the current generated pseudo labels can further boost the student network, then $\lambda_{\mathrm{meta}}$ will be negative, and positive vice versa.\\

\noindent \textbf{Teacher Network} 
The teacher network is jointly optimized by two objectives: a typical semi-supervised learning loss $\mathcal{L}_{\mathrm{ssl}}$ and a meta learning loss $\mathcal{L}_{\mathrm{meta}}$:
\begin{equation*}
\mathcal{L}_{\mathrm{T}}=\mathcal{L}_{\mathrm{ssl}}+\mathcal{L}_{\mathrm{meta}}.
\end{equation*}

\noindent For the SSL loss, it consists of the supervised loss on the strongly-labeled data and the regularization loss on the weakly-labeled data.
\begin{equation*}
\mathcal{L}_{\mathrm{ssl}}=\mathcal{L}_{\mathrm{sup}}+\mathcal{L}_{\mathrm{reg}}.
\end{equation*}

\noindent The supervised loss $\mathcal{L}_{\mathrm{sup}}$ is defined as
\begin{equation*}
\mathcal{L}_{\mathrm{sup}}=\frac{1}{N_{l}}\sum_{i=1}^{N_{l}} \ell(\mathbf{Y}_i^{l}, f(\mathbf{X}_i^{l}; {\bm \theta_{T}}))
\end{equation*}

The regularization loss $\mathcal{L}_{\mathrm{reg}}$ alleviates the overfitting of the teacher by enforcing the prediction consistency between the original and augmented weakly-labeled samples.

\begin{equation*}
\mathcal{L}_{\mathrm{reg}} =-\frac{1}{N_{w}*M}\sum_{i=1}^{N_{w}}\sum_{j=1}^{M} \frac{f(x_{ij}^w; \bm{\theta}_{T})}{ \tau} \log(f(\tilde{x}_{ij}^w; \bm{\theta}_{T})))
\end{equation*}

where $f({x}_{ij}^w; \bm{\theta}_{T})$ is the prediction logits of the teacher network on the $j$-th token of the $i$-th weakly-labeled sample $\mathbf{X}_{i}^w$. $\log(f(\tilde{x}_{ij}^w; \bm{\theta}_{T}))$ is the prediction logits of the corresponding token of the augmented weakly-labeled sample $\tilde{\mathbf{X}}_{i}^w$ and $\tau$ is the temperature factor to control the smoothness. Here, we do not explicitly augment the sentence and instead add random Gaussian noises $G(\mathbf{0}, {\bm \sigma}^{2})$ to the BERT embedding of each token to increase the diversity of the sentence. 

The meta loss $L_{\mathrm{meta}}$ for the teacher network is defined as:
\begin{equation*}
L_{\mathrm{meta}}=\frac{\lambda_{\mathrm{meta}}}{N_{w}}\sum_{i=1}^{N_{w}} \ell(\mathbf{Y}_i^{r}, f(\mathbf{X}_i^{w}; \bm{\theta}_{T}))
\end{equation*}
The performance variation of the student network on the strongly-labeled data is formulated as the feedback signal $\lambda_{\mathrm{meta}}$ to dynamically adapt the teacher network's pseudo-labeling strategies. The teacher and student can have the same encoder (e.g., DistilBERT~\cite{Sanh2019DistilBERTAD}), or a larger teacher for better prediction (e.g., BERT~\cite{devlin2019bert}) and a small student (e.g., DistilBERT) for fast online production inference.

\subsection{QUEACO Attribute Value Normalization}
In this section, we discuss two different types of AVN method for and the product type attribute and general attributes, respectively.

\subsubsection{AVN for Product type attribute}\label{q2pt}

E-commerce websites usually have their own self-defined product category taxonomy, which is used for organizing and indexing the products. Thus, identifying the product type of a given query is one of the most critical components of the query attribute value extraction.

However, there are three challenges in directly normalizing the surface form product type: 1) some queries do not have explicit surface form product type while they are implicitly associated with some product types. For example, as shown in Table~\ref{tab:product_type_case_study} case\#2, there is no surface form product type in a movie query ``{\it wonder woman 1984}'', but the product type of the query is ``{\it movie}''; 2) many entity mentions are the hyponyms of product type values. For example, as shown in Table~\ref{tab:product_type_case_study} case\#6, for the query ``{\it mini pocket detangler brush}'', its surface form product type ``{\it detangler brush}'' is a hyponym of its product type ``{\it hair brush}''; 3) the same surface form might correspond to different product types. For example, the product type of the query ``{\it tote for travel}" is ``{\it luggage}'', but the product type of the query ``{\it mk tote for woman}" is ``{\it handbag}''. 

Alternatively, we can get the query-to-productType associations using the weakly-labeled behavior data. For frequent queries, we use query search logs to get the product type relevance vector $\small{\mathbf{Y}_{i}^{\mathrm{pt}}\!=\!{ P(C_o=\mathbf{V}|\mathbf{X}_{i})}}$ of query $\mathbf{X}_{i}^{w}$ as defined in Section~\ref{avn_notations}, and then get the most relevant product types. Given that not all queries have enough user-behavioral signals, we use this weakly labeled data $D \!=\!\{(\mathbf{X}_{i}^{w}, \mathbf{Y}_{i}^{\mathrm{pt}})\}_{i=1}^{N_w}$ to train a multi-label query classification model~\cite{hashemi2016query,kim2016intent,lin2020light} for predicting the product type distribution of less frequent queries. To meet the latency constraint, we also use DistilmBERT as the encoder.

\begin{table}[htb!]
\centering
\resizebox{1\columnwidth}{!}{
\begin{tabular}{ c| l|c|c   } 
 \toprule
 \hline
    Case\# & Query & surface form & Behavior-based  \\ 
    \hline
1 & nike & None & shoes \\
\hline
2 & wonder woman 1984 & None & movie \\
\hline
3 & unicorn & None & clothes, toys\\
\hline
4 &lg smart tv 32 & smart tv & television \\
\hline
5 & patio umbrella & patio umbrella & umbrella \\
\hline
6 & mini pocket detangler brush & detangler brush & hair brush \\
\hline
7 & tote for travel & tote & luggage \\
\hline
8 & mk tote for women & tote & handbag \\
 \hline
 \bottomrule
\end{tabular}}
\caption{Case study on surface \& behavior-based product type.}
\vspace{-9mm}
\label{tab:product_type_case_study}
\end{table}

\subsubsection{AVN for general attributes}
The attribute value normalization corresponds to the entity disambiguation task in entity linking. Prior entity linking works for search queries~\cite{cornolti2014smaph,tan2017entity,blanco2015fast} leverage additional information, such as knowledge base and query log, and search results. Inspired by this, we propose to extract common surface form to canonical form mapping based on QUEACO NER predictions and weakly-labeled query-to-attribute associations. 

We use the entity type ``brand'' $b$ as the example. Using the method defined in Section~\ref{avn_notations}, we can get the most relevant brand $b_{i}$ for the query $\mathbf{X}^{w}_{i}$ by aggregating the query search logs. Then we can associate surface form brand ${X}_{i}^{w,b}$ and the most relevant behavior-based brand $b_{i}$ through the query $\mathbf{X}^{w}_{i}$. Given a surface form brand value $m$ and a canonical form brand value $v$, we can define the mapping probability between them as,

\begin{equation*}
P(v|m) = \frac{\sum^{N_{w}}_{i} \vmathbb{1}(X_{i}^{w,b}=m, b_{i}=v)}{\sum^{N^{w}}_{i} \vmathbb{1}(X_{i}^{w,b}=m)}.
\end{equation*}

However, we find the same surface form can be normalized to different canonical forms depending on the query context. For example, as shown in Table~\ref{tab:product_type_case_study} case\#1 and \#2, the same surface form size ``{\it apple}'' can be mapped to ``{\it apple barrel}'' given the query ``{\it apple craft paint}'', and ``{\it Apple computer}'' given the query ``{\it apple macbook pro}''. The finding is consistent with the recent embedding-based entity linking works~\cite{wu2020scalable,agarwal2020entity}. However, due to the strict requirement on the inference latency and very high request volume, it is hard to directly apply the current state-of-the-art embedding-based entity disambiguation models, which use the context embedding for the candidate ranking, to the query side~\cite{yamada2016joint,yamada2019global,gillick2019learning,wu2020scalable,agarwal2020entity}. Alternatively, we simplify the setting by using query product type as the context of the query. We then define the probability of one surface form value $m$ conditioned on canonical form attribute value $v$, given product type $p$ as:
\begin{equation*}
P(v|m,p) = \frac{\sum^{N_{w}}_{i} \vmathbb{1}(X_{i}^{w,b}=m, b_{i}=v,  Y_{i}^{\mathrm{pt}}=p)}{\sum^{N_{w}}_{i} \vmathbb{1}(X_{i}^{w,b}=m,Y_{i}^{\mathrm{pt}}=p)}
\end{equation*}

\begin{table}[htb!]
\centering
\resizebox{1\columnwidth}{!}{
\begin{tabular}{ c| l|c|c |c  } 
 \toprule
 \hline
    Case\# & Query & entity & surface form  & canonical form  \\ 
    \hline
1 & lg smart tv 32 & size & 32 & 32 inch \\
\hline
2 & fish tank 32 & size & 32 & 32 gallon \\
\hline
3 & apple craft paint & brand & apple & apple barrel\\
 \hline
4 & apple macbook pro  & brand & apple & Apple computer \\
 \bottomrule
\end{tabular}}
\caption{Case study on surface \& canonical value.}
\label{tab:AVN_case_study}
\vspace{-8mm}
\end{table}

\section{Experiments}

\subsection{QUEACO query NER}

\subsubsection{Data Description}\label{data_statistics}
We collect search queries from a real-world e-commerce website and construct two datasets: (1) strongly-labeled dataset, which is human annotated, and (2) weakly-labeled dataset, which is generated through the partial query tagging, as shown in Table~\ref{tab:ws_example}. 
The statistics of the strongly-labeled and the weakly-labeled datasets are shown in table~\ref{tab:datasets} and table~\ref{tab:weak_datasets}. The details of these datasets are shown below:

\begin{itemize}
\item We split the entire dataset into train/dev/test by $90\%$, $5\%$, and $5\%$. The size of strongly-labeled and the weakly-labeled training data are 677K and 17M. The weakly-labeled dataset is more noisy and is more than 26 times bigger than the strongly-labeled dataset.

\item The strongly-labeled data contains 12 languages: English (En), German (De), Spanish (Es), French (Fr), Italian (It), Japanese (Jp), Chinese (Zh), Czech (Cs), Dutch (Nl), Polish (Pl), Portugal (Pt), Turkish (Tr). The weakly-labeled dataset does not have Zh, CS, NI, and PI languages.

\item The non-O \%coverage for the strongly-labeled dataset is $98.31\%$, and there are 13 non-O types. However, the non-O \%coverage for weakly-labeled data is $43.21\%$, and there are 11 non-O types, indicating the weak labels suffer from severe incompleteness issues. The incomplete annotation is due to the exact string matching between query span and product attribute values~\cite{mehta2021latex}. Table~\ref{tab:weak_datasets} also presents the precision and recall of weak label performance on an evaluation golden set. In particular, the overall recall is lower than 50, which is consistent with the non-O \%coverage. The low recall issue is even more severe for low-resource languages, like Jp, Pt, and Tr. At the same time, the weak labels also suffer from labeling bias since the overall precision is lower than $80\%$.
\end{itemize}

\begin{table}[htb!]
\centering
\resizebox{0.9\columnwidth}{!}{
\fontsize{8.5}{10.2}\selectfont
\begin{tabular}{@{ }c@{ }|@{ }c@{ }|@{ }c@{ }|@{ }c@{ }|@{ }c@{ }|@{ }c@{ }} 
 \toprule
 \hline
    Dataset &\#Train &\#Dev &\#Test &\# Non-O Type & Non-O \%Coverage  \\ 
  \hline
En &256571	&14193	&14269  & 13  & 98.87\\
De &98980	&5442	&5473   & 13  & 95.49\\
Es &63844	&3600	&3488   & 13  & 99.05\\
Fr &79176	&4383	&4504   & 13  & 98.91\\ 
It &52136	&2933	&2867   & 13  & 99.04 \\
Jp &77457	&4422	&4365   & 13  & 98.65   \\
Zh &22467	&1238	&1247   & 13  & 98.51 \\
Cs &4430	&272	&252    & 13  & 93.66  \\
Nl &8562	&423	&478    & 13  & 97.09    \\
Pl &4489	&251	&229    & 13  & 92.19   \\
Pt &4467	&273	&247    & 13  & 99.45  \\
Tr &5093	&267	&274    & 13  & 99.52\\
Total & 677672    &37697    &37693     &13  &98.31 \\
 \hline
 \bottomrule
\end{tabular}}
\caption{The data statistics of strongly-labeled NER dataset. }
\label{tab:datasets}
\vspace{-5mm}
\end{table}

\begin{table}[htb!]
\centering
\resizebox{1.0\columnwidth}{!}{
\fontsize{8.5}{10.2}\selectfont
\begin{tabular}{@{ }c@{ }|@{ }c@{ }|@{ }c@{ }|@{ }c@{ }|@{ }c@{ }|@{ }c@{ }} 
 \toprule
 \hline
    Dataset &\#Train  &\# Type &  \%Coverage & Span Precision & Span Recall \\ 
 \hline
En & 14144225  &  11 &  42.64  & 78.50 & 47.53\\
De & 2004144 &  11 & 48.55  & 83.18 & 52.35\\
Es & 322435  &  11 &   45.79 & 82.24 & 51.32\\
Fr & 504309  &  11 &    49.00 & 81.15 & 51.56 \\
It & 475594  & 11  &   48.87 & 81.69 & 50.82\\
Jp & 241078  &  11 &  20.80 & 67.67 & 25.53\\
Pt & 134458  & 11  &  33.91  & 80.83 & 32.23\\
Tr & 23980 &  11 &  32.87 & 86.12 & 34.95\\
Total & 17850787    &11    &43.21 & 79.80 & 48.04\\ 
 \hline
 \bottomrule
\end{tabular}}
\caption{The data statistics of weakly-labeled NER dataset. Type and Coverage denote the number of entity type and the ratio of non-\texttt{O} entity.}
\label{tab:weak_datasets}
\vspace{-8mm}
\end{table}

\subsubsection{Evaluation Metrics}
We use the span-level micro precision, recall and F1-score as the evaluation metrics for all experiments. For the per language experiment, we only report the span-level micro F1-score for each language, due to the space limit.

\subsubsection{Analysis of the Base Encoder}\label{discuss_base_encoder}

We benchmark the DistilmBERT performance with the baseline models in the query attribute extraction literature. All RNN experiments use FastText multi-lingual word embeddings \cite{conneau2017word} and the TARGER implementation \cite{chernodub2019targer}.
\begin{itemize}
    \item  RNN models: BiLSTM, BiGRU, BiLSTM-CRF and BiGRU-CRF models are benchmarked for the Home Depot query NER model~\cite{cheng2020end}.
    \item BiLSTM-CNN-CRF~\cite{lample2016neural,ma2016end} is the state-of-the-art NER model architecture before BERT~\cite{devlin2019bert,yangxlnet}.
    \item DistilmBERT baselines: 1) DistilmBERT (Single) means separately finetuning DistilmBERT on the strongly-labeled data for each single language. 2) DistilmBERT (Multi) means finetuning DistilmBERT on the strongly-labeled data for all languages.
\end{itemize}  

\begin{table}[htb!]
\centering
\resizebox{0.7\columnwidth}{!}{
\begin{tabular}{@{ }c@{ }|@{ }c@{ }|@{ }c@{ }|@{ }c@{ }|@{ }c@{ }|@{ }c@{ }} 
 \toprule
 \hline
    Method (\textit{Span level}) & Precision & Recall & F1 \\ 
 \hline
 \hline
 BiLSTM   & 65.66 & 70.09 & 67.81
 \\
 BiGRU  &  64.35 &	68.96 & 66.58
 \\
 BiLSTM-CRF  & 71.04 & 69.36 & 70.19\\
 BiGRU-CRF  & 69.45 & 67.98 & 68.71\\
 BiLSTM-CNN-CRF  & 70.33 & 67.92 & 69.11\\
 BiGRU-CNN-CRF & 67.75 & 65.40 & 66.56\\
 DistilmBERT (Single) & 71.72 & 74.16 & 72.92  \\
 DistilmBERT (Multi)  & 73.33 & 75.29 & 74.29 \\
 \hline
 \bottomrule
\end{tabular}}
\caption{Comparison of different encoders.}
\label{tab:baseline_ner_results}
\vspace{-8mm}
\end{table}

As shown in Table~\ref{tab:baseline_ner_results}: the DistilmBERT has better performance than other non-BERT baselines. Furthermore, finetuning DistilmBERT with all languages has better performance than training a separate model for each language.

\subsubsection{Discussion on the training data} \label{discuss_training_data}

\begin{table*}[t]
\begin{tabular}{ccc}
\includegraphics[width=0.32\textwidth]{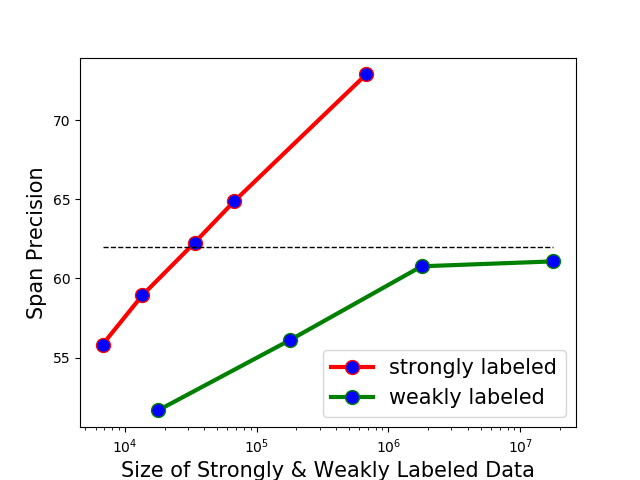} &
\includegraphics[width=0.32\textwidth]{figure/precision.png} &
\includegraphics[width=0.32\textwidth]{figure/precision.png} \\
 (a) \textbf{Span Precision} & (b) \textbf{Span Recall} & (c) \textbf{Span F1} \
 \end{tabular}
 \vspace{-2mm}
\captionof{figure}{Size of strongly \& weakly labeled data vs. Performance. All results are produced by directly finetuning the DistilmBERT model with the subsampled dataset. We subsample $1\%$, $2\%$, $5\%$, $10\%$ and $100\%$ of the 677K strongly-labeled data, and subsampled  $0.1\%$, $1\%$, $10\%$ and $100\%$ of the 17M weakly-labeled data. In (a), (b) and (c), span-level precision, recall and micro-f1 are shown.}
\label{fig:perf_vs_size}
\vspace{-5mm}
\end{table*}


In this section, we discuss the use of training data for QUEACO query NER model. We benchmark our setting with the baseline in the query NER literature, where only weakly-labeled data is available. All experiments use DistilmBERT as the base NER model for a fair comparison.

In Figure~\ref{fig:perf_vs_size}, we subsample the strongly and weakly-labeled dataset and we find:
\begin{itemize}
    \item The precision and recall of the model trained with the weakly-labeled data do not change much when the training data size increases from $10\%$ to $100\%$. However, both the precision and recall increase dramatically when size of strongly-labeled data increases, especially the precision.
    \item The best precision that weakly-labeled data can achieve is around $60\%$. However, 34K strongly-labeled queries can already achieve $62.27\%$ precision. And the precision reaches to $72.90\%$ when trained with 677K strongly-labeled queries. With only weakly-labeled data, the best recall is only around $26\%$, much lower than that using strongly-labeled data. 7K strong-labeled data can already achieve $48.78\%$ recall. 
\end{itemize}

The findings are consistent with the conclusion of BOND~\cite{liang2020bond} that pre-trained language models can easily overfit to incomplete weak labels. And this explains why the existing query NER works~\cite{cheng2020end,wen2019building,cowan2015named,kozareva2016recognizing} do not adopt the state-of-the-art pre-trained language model.

\begin{figure}[htb!]
    \centering
    \includegraphics[width=0.32\textwidth]{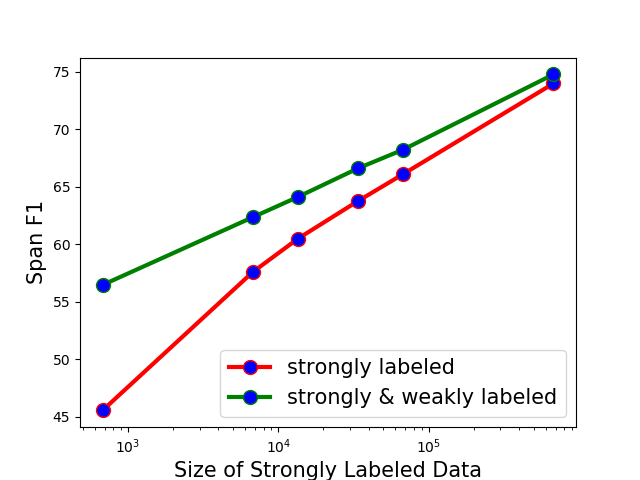}
    \caption{Size of Strongly Labeled Data vs. Micro span-level F1. "strongly labeled": a baseline that finetunes DistilmBERT with the strongly labeled data, "strongly \& weakly labeled":  a baseline that pretrains Distil-mBERT with weakly labels and then finetunes it on the strongly labeled data.}
\label{fig:perf_vs_strong_size}
\vspace{-5mm}
\end{figure}

In Figure~\ref{fig:perf_vs_strong_size}, we show the performance improvement for introducing weakly-labeled data to different sizes of randomly sub-sampled strongly-labeled data. It is shown that the smaller strongly-labeled data, the bigger improvement the weak labels can introduce. However, the performance improvement is marginal when the strongly-labeled dataset is sufficient. In section~\ref{query_ner_model}, we introduce the QUEACO query NER model to better utilize the weak labels to further improve the query NER model performance.

\subsubsection{Implementation Details of QUEACO}

We employ the DistilmBERT~\cite{Sanh2019DistilBERTAD} with 6 layers, 768 dimension, 12 heads and 134M parameters as our encoder. We use ADAM optimizer with a learning rate of $10^{-5}$, tuned amongst \{$10^{-5}$, $2\times10^{-5}$, $3\times10^{-5}$, $5\times10^{-5}$, $10^{-4}$\}. We search the number of epochs in [1,2,3,4,5] and batch size in [8, 16, 32, 64]. The Gaussian noise variance $\bm{\sigma}$ is tuned amongst \{0.01, 0.1, 1.0\}. The temperature factor for smoothness $\tau$ is tuned amongst \{0.5, 0.6, 0.7, 0.8, 0.9\}. The threshold ${\epsilon}$ is tuned amongst \{0.5, 0.6, 0.7, 0.8, 0.9\}. All implementations are based on transformers in Pytorch 1.7.0. To alleviate overfitting, we perform early stopping on the validation set during both the pretraining and finetuning stages. For model training, we use an Amazon EC2 virtual machine with 8 NVIDIA A100-SXM4-40GB GPUs, configured with CUDA 11.0.

\subsubsection{Baseline Models}
As discussed in section~\ref{model_architecture_benchmark} and section~\ref{discuss_training_data}, it is evident that the setting of using DistilmBERT as base NER model and using both strongly and weakly-labeled dataset as training data, outperforms the other settings. We also conduct baseline experiments in similar settings to show the effectiveness of the QUEACO query NER model. All experiments use DistilmBERT as the base NER model for the fair comparison.

\noindent~$\bullet$ \textbf{Supervised Learning Baseline}: We directly fine-tune the pre-trained model on the strongly-labeled data.

\noindent~$\bullet$ \textbf{Semi-supervised Baseline}

\begin{itemize}
    \item Self Training: self-training with hard pseudo-labels
    \item NoisyStudent~\cite{xie2020self} extends the idea of self-training and distillation with the use of noise added to the student during learning. 
\end{itemize}

\noindent~$\bullet$ \textbf{Weakly-supervised Baseline}: Similar to QUEACO, these weakly-supervised baselines also have two stages: pretraining with strongly-labeled and weakly-labeled data, and finetuning with strongly-labeled data. We only report stage 2 performance.

\begin{itemize}
    \item Weakly Supervised Learning (WSL): Simply combining strongly-labeled data with weakly-labeled data~\cite{mann2010generalized}.
    
    \item Weighted Weakly Supervised Learning (Weighted WSL): WSL with weighted loss, where weakly-labeled samples have a fixed smaller weight and strongly-labeled samples have weight = 1.  We tune the weight and present the best result. 
 
    \item Robust WSL: WSL with mean squared error loss function, which is robust to label noises~\cite{ghosh2017robust}. 
    
    \item BOND (hard/soft): BOND~\cite{liang2020bond} employs a state-of-the-art two-stage teacher-student framework with hard pseudo-labels or soft pseudo-labels~\cite{xie2016unsupervised}.
    
    \item BOND (soft-high): only uses the soft pseudo-labels, with high confidence selection for student network training in the BOND framework.
    
    \item BOND (NoisyStudent): applies noisy student~\cite{xie2020self} to the BOND framework.
\end{itemize}

\subsubsection{Main results} From Table ~\ref{tab:semi_supervised_ner_results} and ~\ref{tab:per_language_result}, our results obviously demonstrate the effectiveness of our proposed QUEACO query NER model:
\begin{itemize}
    \item The proposed QUEACO query NER model achieves the state-of-the-art performance. More specifically, we can improve upon the best weakly-supervised baseline model by a margin of $0.4\%$ on micro span-level F1. QUEACO query NER model with mBERT as the teacher network can further enhance the model performance.
    \item We also find weak labels improve by $1.09\%$ w.r.t the best semi-supervised result, showing the weak labels have useful information if utilized effectively.
    \item Table~\ref{tab:per_language_result} compares the span F1 between the baseline DistilmBERT model and the QUEACO query NER model for each language. We can observe consistent performance improvement for the high resource languages (En, De, Es, Fr, It, Jp).  On the other hand, we observe performance drop for those low resources languages with a few or no weakly-supervised data (Cs, Nl, Pl, Tr). Pt is also a low-resource language but observes significant performance improvement because we have more than 100k weakly supervised training data for Pt. We believe we can further improve the performance of those low-resource languages if more weak supervised data is collected.
\end{itemize}

\begin{table}[t]
\centering
\resizebox{0.9\columnwidth}{!}{
\begin{tabular}{ l | ccc   } 
 \toprule
 \hline
    Method (\textit{Span level}) & Precision & Recall & F1 \\ 
 \hline
\multicolumn{4}{c}{Supervised Baseline}\\
\hline
 DistilmBERT (Single) & 71.72 & 74.16 & 72.92 \\
 DistilmBERT (Multi) & 73.33 &	75.29 &	74.29 \\
 \hline
 \hline
 \multicolumn{4}{c}{Semi-supervised Baseline (Encoder: DistilmBERT)}\\
 \hline
ST   & 73.29	& 75.44 &74.35 \\
Noisy student &  73.28 & 75.38& 74.32	 \\
 \hline
 \hline
 \multicolumn{4}{c}{Weakly-supervised Baseline (Encoder: DistilmBERT)}\\
 \hline
unweighted WSL  & 73.81 &  75.93 & 74.85 \\
 weighted WSL  & 73.77 & 75.97 & 74.85\\
 robust WSL & 73.10 & 75.20 & 74.14\\
 BOND hard & 73.77 &	 75.81	& 74.78 \\
 BOND soft & 73.65 & 75.68 &	74.65 \\
 BOND soft high conf & 73.95	& \textbf{76.05}	& \textbf{74.98} \\
 BOND noisy student & \textbf{73.97} & 75.99 & 74.97 \\
 \hline
  \multicolumn{4}{c}{Ours (Student: distillmBERT)}\\
 \hline
QUEACO  (Teacher: distilmBERT) & 74.44 & 76.35 & 75.38\\
QUEACO  (Teacher: mBERT) & \textbf{74.48} & \textbf{76.41} & \textbf{75.44}\\
$\Delta$ & (\textcolor{green}{\textbf{+0.51}})  & (\textcolor{green}{\textbf{+0.36}}) & (\textcolor{green}{\textbf{+0.46}}) \\
 \hline
 \bottomrule
\end{tabular}}
\caption{Comparison between QUEACO and baseline methods on micro span-level F1.}
\label{tab:semi_supervised_ner_results}
\vspace{-8mm}
\end{table}

\begin{table}[t]
\centering
\resizebox{1\columnwidth}{!}{
\begin{tabular}{ l | ccc  } 
 \toprule
 \hline
    Language  & Weakly Data available & DistilmBERT (Multi) & QUEACO  \\ 
    \hline
En & True & 75.42 & 76.97 (\textcolor{green}{\textbf{+1.55}})\\
De & True & 75.26 & 76.70 (\textcolor{green}{\textbf{+1.44}})\\
Es & True & 77.30 & 77.67 (\textcolor{green}{\textbf{+0.37}})\\
Fr & True & 71.56 & 73.20 (\textcolor{green}{\textbf{+1.64}})\\ 
It & True & 77.88 & 78.42 (\textcolor{green}{\textbf{+0.54}})\\
Jp & True & 65.49 &  65.88 (\textcolor{green}{\textbf{+0.39}})\\
Zh & False & 71.02 & 72.19 (\textcolor{green}{\textbf{+1.17}})\\
Cs & False & 72.61 & 70.93 (\textcolor{red}{\textbf{-1.68}})\\
Nl & False & 75.46 &  75.30 (\textcolor{red}{\textbf{-0.16}})\\
Pl & False &  79.71 & 79.43 (\textcolor{red}{\textbf{-0.28}})\\
Pt & True &  58.24 & 62.00 (\textcolor{green}{\textbf{+3.76}})\\
Tr & True &  72.12 & 71.80 (\textcolor{red}{\textbf{-0.32}})\\
 \hline
 \bottomrule
\end{tabular}}
\caption{Comparison between DistilmBERT (Multi) and QUEACO for each language on micro span-level F1.}
\label{tab:per_language_result}
\vspace{-8mm}
\end{table}

\subsubsection{Ablation Study}
\begin{itemize}
    \item QUEACO w/o student feedback $\mathcal{L}_\text{meta}$: use a fixed teacher network to generate pseudo labels for a student network.
    \item QUEACO w/o noise: remove random Gaussian noise added to the BERT embedding when training the teacher network.
    \item QUEACO w/o weak labels: remove the pseudo \& weak label refinement step, and only use the pseudo labels for student network training.  
    \item QUEACO w/o finetune: remove stage 2: strong labels finetuning.
\end{itemize}

As shown in table~\ref{tab:ablation_study}, we find the final finetuning is essential to QUEACO NER. All components from QUEACO, including student feedback, random Gaussian noise to the BERT embedding and the pseudo \& weak label refinement, are effective. 

\begin{table}[htb!]
\centering
\resizebox{1\columnwidth}{!}{
\begin{tabular}{ l | ccc   } 
 \toprule
 \hline
    Method (\textit{Span level}) & Precision & Recall & F1 \\ 
\hline
QUEACO w/o student feedback &  74.09 &  76.11 & 75.09\\
QUEACO w/o noise &  74.18 &  76.01 & 75.08\\
QUEACO w/o weak labels & 74.04  & 75.77 & 74.89\\
QUEACO w/o finetune &  63.31 & 66.62 & 64.92\\
QUEACO & 74.44 & 76.35 & 75.38\\
 \hline
 \bottomrule
\end{tabular}}
\caption{Ablation study.}
\label{tab:ablation_study}
\vspace{-8mm}
\end{table}

\subsection{QUEACO Attribute Value Normalization}
\subsubsection{Product type AVN}
In the query NER, the span-level micro F1-score for product type is only $77.12\%$. The performance for NER-based product type value extraction will be even worse, since many surface forms cannot be normalized. In Table~\ref{tab:q2pt}, we show the product type precision, recall and F1 of the multi-label query classification model, as described in section~\ref{q2pt}, on a golden set. We can conclude the query classification approach, trained with weakly-labeled data, is more suitable to product type attribute extraction than query NER. 

\begin{table}[htb!]
\centering
\resizebox{0.9\columnwidth}{!}{
\begin{tabular}{ l | cccc   } 
 \toprule
 \hline
    Country & Eval Data Size & Precision & Recall & F1 \\ 
\hline
USA &  2746 &  85.13 & 81.1 & 83.08\\
UK &  2590 &  85.44 &	85.71 & 85.58\\
Canada & 2705  & 85.07 & 86.41 & 85.73\\
Japan  &  2151 & 85.2	& 80.06 & 82.55 \\
Germany & 2254 & 85.01 & 88.54 & 86.74\\
 \hline
 \bottomrule
\end{tabular}}
\caption{Product type attribute value extraction performance.}
\label{tab:q2pt}
\vspace{-8mm}
\end{table}

\subsubsection{AVN for other attributes}

In Table ~\ref{tab:attribute_normalization}, we show some attribute normalization result for brand, color and size attributes, using our proposed method. We can see that our proposed method is effective in finding common surface attributes, including: 
\begin{itemize}
    \item spelling error: brand ``{\it Michael Kors}'' is often misspelled as ``{\it Micheal Kors}'', ``{\it Levi's}'' is often misspelled as ``{\it levi}'';
    \item spelling invariants: for example, ``{\it 3 by 5}'' and ``{\it3x5}'' are different variants with the same meaning.
    \item abbreviation: for example, ``{\it mk}'' is the abbreviation for ``{\it Micheal Kors}'', ``{\it wd}'' is the abbreviation for ``{\it Western Digital}'', ``{\it in}'' in the mention ``{\it 8 in}'' is the abbreviation for unit ``{\it inches}''.
\end{itemize}

\begin{table}[htb!]
  \centering
  \resizebox{0.8\columnwidth}{!}{
 \begin{tabular}{c c c c }
 \hline
 attribute & surface form  & product type & canonical form  \\ 
 \hline\hline
 size & 3 by 5 & rug & 3x5\\
 size &	2 pack & air filter &  Value Pack (2)\\
 size & 28 foot	& ladder & 28 Feet\\
size  & 10.5 inch & screen protector & 10.5 Inches \\
size & 8 in & toy figure & 8 inches\\
 \hline
 color & golden & 	belt & Gold\\
 color & turquoise & dress & blue \\
 color & navy blue & dress & blue \\
 \hline
 brand & levi & underpants & Levi's \\
brand & mk	& watch	& Michael Kors \\ 
brand & Micheal Kors & watch & Michael Kors \\
brand & wd	& computer drive	& Western Digital \\
 \hline
\end{tabular}}
\caption{QUEACO attribute normalization result.}
\label{tab:attribute_normalization}
\vspace{-8mm}
\end{table}

\section{QUEACO Online Deployment}
\subsection{Online End-to-End Evaluation}
We conducted an end-to-end evaluation of QUEACO on real-world search traffic. We have two evaluation metrics: span-level precision and token-level coverage.
For span-level precision, we resort to a crowdsourcing data labeling platform called Toloka\footnote{https://toloka.yandex.com} and the reported overall precision of the QUEACO system is $97\%$. Since the query attribute value extraction is an open-domain problem, the human annotators cannot verify the recall of the extracted attribute spans. Therefore, we calculate token-level coverage, i.e., the percentage of tokens annotated by QUEACO, as an approximation of recall. The token-level coverage increased by $38.2\%$ compared to the current system.

\subsection{Application: Extracted Attribute Value for Product Reranking}
To validate the effectiveness of QUEACO signal on the product search system, we design a downstream task, \textit{product reranking}, whose goal is to rerank the top-16 products based on their relevance to the query intent. Specifically, we first use QUEACO to extract attributes for the product search queries. Then, we generate boolean features, such as \textit{is pt match}, \textit{is brand match}, based on the attribute values of queries and products. We refer to these boolean features as QUEACO features. We then train two learning-to-rank (LTR) models: one model uses QUEACO features while the other does not. All other features, settings and hyperparameters of these two models are the same. To compare these two models, we use NDCG@16, which is the normalized discounted cumulative gain (NDCG) score for the top 16 products of the search result. We conducted online A/B experiments for this reranking application in four countries: India, Canada, Japan, and Germany. On average, we improve the NDCG@16 by $0.36\%$.




\section{Related Work}
\subsection{E-commerce Attribute Value Extraction}

Most of the previous works on e-commerce attribute value extraction focus on extracting surface-form attribute values from product titles and descriptions. Some early machine learning works formulate the task as a (semi-) classification problem~\cite{ghani2006text,probst2007semi}. Later, several researchers~\cite{putthividhya2011bootstrapped,more2016attribute} employ a sequence tagging formulation and adopt the CRF model architecture. With the recent advances in deep learning, many RNN-CRF based models are applied to the sequence tagging task~\cite{huang2015bidirectional,lample2016neural,ma2016end}, and have achieved promising results. Following this trend, recent works on the product attribute value extraction task~\cite{zheng2018opentag,xu2019scaling,mehta2021latex} also adopt variants of the BiLSTM-CRF model architecture. In addition, some recent studies have explored BERT-based~\cite{devlin2019bert} Machine Reading Comprehension (MRC)~\cite{xu2019scaling} and Question \& Answering (Q\&A)~\cite{wang2020learning} formulation.

Query attribute value extraction works~\cite{cheng2020end,wen2019building,cowan2015named,kozareva2016recognizing} also employ the sequence tagging formulation and adopt BiLSTM-CRF model architectures as well as its variants. Recent works~\cite{cheng2020end,wen2019building} utilize large behavioral-based data to generate partial query tagging as distant supervision to train the NER model, and they also explore data augmentation and active learning to deal with the data quality issues.

\subsection{NER with Distant Supervision}
To alleviate human labeling efforts, various approaches such as transfer learning~\cite{pan2009survey}, semi-supervised learning~\cite{chapelle2009semi}, and weakly-supervised learning~\cite{zhou2018brief} are emerging and widely applied to low-resource NLP tasks~\cite{zhang2020data,li2020learn,liu2021improving}, e.g., sentiment classification~\cite{li2017end,li2018hierarchical,li2019exploiting,li2019sal}, information extraction~\cite{he2017autoentity,shang2018automated,li2020unsupervised}, etc. Specifically, distant supervision is a type of weak supervision, and is automatically generated based on some heuristics, such as matching spans of unlabeled text to a domain dictionary~\cite{shang2018learning,liang2020bond}. Existing works on NER with distant supervision~\cite{shang2018learning,liang2020bond} mainly focus on the setting that can only access distant supervision. Besides, most existing query NER works~\cite{cheng2020end,wen2019building} only rely on the distant supervision, generated from partial query tagging, for NER model training.

However, in some cases both strongly-labeled data and a large amount of distant supervision are available. The strongly-labeled data, though expensive to collect, is validated to be critical to boost distant supervised NER performance~\cite{Jiang2021NamedER}.
\section{Conclusion}
This paper proposes to utilize the weakly-labeled behavioral data to improve the named entity recognition and attribute value normalization phases of query attribute value extraction. We conduct extensive experiments on a real-world large-scale E-commerce dataset and demonstrate that the QUEACO NER can achieve the state-of-the-art performance and the QUEACO AVN effectively normalizes some common customer typed surface forms. We also validate the effectiveness of the proposed QUEACO system for the downstream product reranking application. 

\bibliographystyle{ACM-Reference-Format}
\bibliography{ref}

\end{document}